\documentclass{article}
\usepackage{amsmath}
\usepackage{amssymb}
\usepackage{graphicx}
\usepackage{color}
\usepackage{url}
\usepackage{amsfonts, amscd, amsthm}
\usepackage{subfigure}
\usepackage{bbm}

\newcommand{\eps}{\varepsilon}

\newcommand{\rank}{\mathrm{rank}}
\newcommand{\<}{\langle}
\renewcommand{\>}{\rangle}

\newcommand{\tensor}{\otimes}

\renewcommand{\mathbf}{\boldsymbol}

\renewcommand{\P}{\mathbb{P}}

\newcommand{\downto}{\searrow}

\newcommand{\bb}{\mathbb}
\newcommand{\mb}{\mathbf}
\newcommand{\mc}{\mathcal}
\newcommand{\mf}{\mathfrak}

\newcommand{\prob}[2][]{\P_{#1}\left[ \, #2 \, \right] }

\newcommand{\expect}[2][]{{\bb E}_{#1}\left[ #2 \right]}

\newcommand{\norm}[2]{\left\| #1 \right\|_{#2}}
\newcommand{\innerprod}[2]{\left\< #1, #2 \right\>}
\newcommand{\magnitude}[1]{\left|#1\right|}

\newcommand{\set}[1]{\left\{ #1 \right\}}

\newcommand{\reals}{\mathbb{R}}

\newcommand{\nullspace}[1]{\mathrm{null}\left( #1 \right)}

\newcommand{\indicator}[1]{\mathbbm{1}_{#1}}
\newcommand{\semidefinitecone}[1]{\mc S^n_+}

\newcommand{\bd}{\boldsymbol}
\newcommand{\mcb}[1]{{\bd{\mc{#1}}}}

\newtheorem{theorem}{Theorem}
\newtheorem*{example*}{Example}
\numberwithin{equation}{section}

\newtheorem{corollary}[theorem]{Corollary}

\newtheorem{definition}{Definition}

\newtheorem{lemma}{Lemma}

\newcommand{\tensorX}{\mb{\mathcal{X}}}
\newcommand{\tensorW}{\mb{\mathcal{W}}}

\newcommand{\tensorXo}{{{\mb{\mathcal{X}}}_0}}
\newcommand{\br}{\mathbb{R}}

\renewcommand{\rank}{\mathrm{rank}}

\newcommand{\tensorY}{\boldsymbol{\mathcal{Y}}}

\newcommand{\tensorZ}{\boldsymbol{\mathcal{Z}}}

\newcommand{\A}{\mathcal{A}}
\newcommand{\Proj}{\mathcal{P}}

\newcommand{\X}{\mathcal{X}}
\newcommand{\Y}{\mathcal{Y}}

\numberwithin{equation}{section}
\usepackage{amsfonts}
\usepackage{amsmath}
\usepackage[ruled,vlined,linesnumbered]{algorithm2e}
\usepackage{verbatim}
\usepackage{setspace}
\usepackage[abs]{overpic}
\usepackage{graphicx}

\begin{document}

\title{Square Deal: Lower Bounds and Improved Relaxations for Tensor Recovery}
\author{Cun Mu$^1$, Bo Huang$^1$, John Wright$^2$, Donald Goldfarb$^1$ \\
$^1$Department of Industrial Engineering and Operations Research, Columbia University\\
$^2$Department of Electrical Engineering, Columbia University}
\maketitle

\begin{abstract}
Recovering a low-rank tensor from incomplete information is a recurring problem in signal processing and machine learning. The most popular convex relaxation of this problem minimizes the sum of the nuclear norms of the unfoldings of the tensor. We show that this approach can be substantially suboptimal: reliably recovering a $K$-way tensor of length $n$ and Tucker rank $r$ from Gaussian measurements requires $\Omega( r n^{K-1} )$ observations. In contrast, a certain (intractable) nonconvex formulation needs only $O(r^K + nrK)$ observations. We introduce a very simple, new convex relaxation, which partially bridges this gap. Our new formulation succeeds with $O(r^{\lfloor K/2 \rfloor}n^{\lceil K/2 \rceil})$ observations. While these results pertain to Gaussian measurements, simulations strongly suggest that the new norm also outperforms the sum of nuclear norms for tensor completion from a random subset of entries.

Our lower bound for the sum-of-nuclear-norms model follows from a new result on recovering signals with multiple sparse structures (e.g. sparse, low rank), which perhaps surprisingly demonstrates the significant suboptimality of the commonly used recovery approach via minimizing the sum of individual sparsity inducing norms (e.g. $l_1$, nuclear norm). Our new formulation for low-rank tensor recovery however opens the possibility in reducing the sample complexity by exploiting several structures jointly.
\end{abstract}

\section{Introduction}
Tensors arise naturally in problems where the goal is to estimate a multi-dimensional object whose entries are indexed by several continuous or discrete variables. For example, a video is indexed by two spatial variables and one temporal variable; a hyperspectral datacube is indexed by two spatial variables and a frequency/wavelength variable. While tensors often reside in extremely high-dimensional data spaces, in many applications, the tensor of interest is {\em low-rank}, or approximately so \cite{kolda2009tensor}, and hence has much lower-dimensional structure. The general problem of estimating a low-rank tensor has applications in many different areas, both theoretical and applied: e.g.,  estimating latent variable graphical models \cite{Anandkumar2012-preprint}, classifying audio \cite{Mesgarani2006}, mining text \cite{Cohen2012-NIPS}, processing radar signals \cite{Sidiropoulos2010-TSP}, to name a few.

In most part of the paper, we consider the problem of recovering a $K$-way tensor $\mcb X \in \reals^{n_1 \times n_2 \times \cdots \times n_K}$ from linear measurements $\mb z = \mc G[ \mcb{X} ] \in \reals^m$.  Typically, $m \ll N = \prod_{i=1}^K n_i$, and so the problem of recovering $\mcb{X}$ from $\mb z$ is {\em ill-posed}. In the past few years, tremendous progress has been made in understanding how to exploit structural assumptions such as sparsity for vectors \cite{Candes2006-CPAM} or low-rankness for matrices \cite{recht2010guaranteed} to develop computationally tractable methods for tackling ill-posed inverse problems. In many situations, convex optimization can estimate a structured object from near-minimal sets of observations \cite{Negahban2012-StatSci,chandrasekaran2012convex,amelunxen2013living}. For example, an $n \times n$ matrix of rank $r$ can, with high probability, be exactly recovered from  $Cnr$ generic linear measurements, by minimizing the nuclear norm $\norm{\mb X}{*} = \sum_i \sigma_i(\mb X)$. Since a generic rank $r$ matrix has $r ( 2n - r )$ degrees of freedom, this is nearly optimal.

In contrast, the correct generalization of these results to low-rank tensors is not obvious. The numerical algebra of tensors is fraught with hardness results \cite{Hillar2009-pp}. For example, even computing a tensor's (CP) rank,
\begin{equation}
\rank_{\mathrm{cp}}(\tensorX)=\min\left\{ r \;|\; \tensorX = \sum_{i=1}^r \mb a_1^{(i)}\circ \mb a_2^{(i)}\circ \cdots \circ \mb a_K^{(i)} \right\},
\end{equation}
is NP-hard in general. The nuclear norm of a tensor is also intractable, and so we cannot simply follow the formula that has worked for vectors and matrices.

With an eye towards numerical computation, many researchers have studied how to estimate or recover tensors of small {\em Tucker rank} \cite{Tucker1966}.
The Tucker rank of a $K$-way tensor $\mcb{X}$ is a $K$-dimensional vector whose $i$-th entry is the (matrix) rank of the mode-$i$ unfolding $\mcb{X}_{(i)}$ of $\mcb{X}$:
\begin{equation} \label{eqn:tucker-def}
\rank_{\mathrm{tc}}(\tensorX):=\big{(}\rank(\tensorX_{(1)}),\rank(\tensorX_{(2)}),\cdots,\rank(\tensorX_{(K)})\big{)}.\vspace{-.5mm}
\end{equation}
Here, the matrix  $\mcb X_{(i)} \in \reals^{n_i \times \prod_{j\neq i} n_j}$ is obtained by concatenating all the mode-$i$ fibers of $\bd{\mc{X}}$ as column vectors. Each {\em mode-$i$ fiber} is an $n_i$-dimensional vector obtained by fixing every index of $\mcb X$ but the $i$-th one. The Tucker rank of $\tensorX$ can be computed efficiently using the (matrix) singular value decomposition. For this reason, we focus on tensors of low Tucker rank. However, we will see that our proposed regularization strategy also automatically adapts to recover tensors of low CP rank, with some reduction in the required number of measurements.

The definition \eqref{eqn:tucker-def} suggests a very natural, tractable convex approach to recovering low-rank tensors: seek the $\mcb{X}$ that minimizes $\sum_i \lambda_i \norm{\mcb{X}_{(i)}}{*}$ out of all $\mcb{X}$ satisfying $\mc{G}[\mcb{X}] = \mb z$. We will refer to this as the {\em sum-of-nuclear-norms (SNN)} model. Originally, proposed in \cite{liu2009tensor}, this approach has been widely studied \cite{gandy2011tensor,signoretto2010nuclear,tomioka2010estimation,tomioka2011statistical,signoretto2013-ML} and applied to various datasets in imaging \cite{Signoretto2011-SPL,Semerci2013-TIP,Kreimer2013-ICASSP,Li2010-ICIP,Li2010-ECCV}. 

Perhaps surprisingly, we show that this natural approach can be {\em substantially suboptimal}, and introduce a simple new convex regularizer with provably better performance. For ease of stating results, suppose that $n_1 = \dots = n_K = n$, and $\rank_{\mathrm{tc}}(\mcb X) \preceq (r,r,\cdots,r)$. Let $\mf T_r$ denote the set of all such tensors. We will consider the problem of estimating an element $\mcb{X}_0$ of $\mf T_r$ from Gaussian measurements $\mc{G}$  (i.e., $\mb z_i = \< \mcb{G}_i, \mcb{X} \>$, where $\mcb{G}_i$ has i.i.d.\ standard normal entries). To describe a generic tensor in $\mf T_r$, we need at most $r^K+rnK$ parameters. Section \ref{sec:nonconvex} shows that a certain nonconvex strategy can recover all $\mcb{X} \in \mf T_r$ exactly when $m > (2r)^K + 2nrK$. In contrast, the best known theoretical guarantee for SNN minimization, due to Tomioka et.\ al.\ \cite{tomioka2011statistical}, shows that $\mcb{X}_0 \in \mf T_r$ can be recovered (or accurately estimated) from Gaussian measurements $\mc G$, provided $m = \Omega( r n^{K-1} )$. In Section \ref{sec:SNN}, we prove that this number of measurements is also {\em necessary}: accurate recovery is unlikely unless $m = \Omega( r n^{K-1})$. Thus, there is a substantial gap between an ideal nonconvex approach and the best known tractable surrogate. In Section \ref{sec:square}, we introduce a simple alternative, which we call the {\em square norm} model, which reduces the required number of measurements to $O(r^{\lfloor K/2 \rfloor}n^{\lceil K/2 \rceil})$. For $K > 3$, this improves by a multiplicative factor polynomial in $n$.

Our theoretical results pertain to Gaussian operators $\mc{G}$. The motivation for studying Gaussian measurements is twofold. First, Gaussian measurements may be of interest for compressed sensing recovery \cite{donoho2006compressed}, either directly as a measurement strategy, or indirectly due to universality phenomena \cite{Donoho2009-TRSA,Bayati2012}. Second, the available theoretical tools for Gaussian measurements are very sharp, allowing us to rigorously investigate the efficacy of various regularization schemes, and prove both upper and lower bonds on the number of observations required. In simulation, our qualitative conclusions carry over to more realistic measurement models, such as random subsampling \cite{liu2009tensor} (see Section \ref{sec:simulation}). We expect our results to be of interest for a wide range of problems in tensor completion \cite{liu2009tensor}, robust tensor recovery / decomposition \cite{Li2010-ECCV,Qin2012-pp} and sensing.

Our technical methodology draws on, and enriches, the literature on general structured model recovery. The surprisingly poor behavior of the SNN model is an example of a phenomenon first discovered by Oymak et.\ al.\ \cite{oymak2012simultaneously}: for recovering objects with multiple structures, a combination of structure-inducing norms is often not significantly more powerful than the best individual structure-inducing norm. Our lower bound for the SNN model follows from a general result of this nature, which we prove using the geometric framework of \cite{amelunxen2013living}. Compared to \cite{oymak2012simultaneously}, our result pertains to a more general family of regularizers, and gives sharper constants. In addition, we demonstrate the possibility to reduce the number of generic measurements through a new convex regularizer that exploit several sparse structures jointly.

\section{Bounds for Non-Convex Recovery}  \label{sec:nonconvex}
\newcommand{\ranktc}[1]{\mathrm{rank}_{\mathrm{tc}}\left( #1 \right) }

In this section, we introduce a non-convex model for tensor recovery, and show that it recovers low-rank tensors from near-minimal numbers of measurements. While our nonconvex formulation is computationally intractable, it gives a baseline for evaluating tractable (convex) approaches.

For a tensor of low Tucker rank, the matrix unfolding along each mode is low-rank. Suppose we observe $\mc G[\mcb{X}_0] \in \reals^m$. We would like to attempt to recover $\mcb{X}_0$ by minimizing some combination of the ranks of the unfoldings, over all tensors $\mcb{X}$ that are consistent with our observations. This suggests a {\em vector optimization} problem  \cite[Chap. 4.7]{boyd2004convex}:
\begin{eqnarray}
\text{\rm minimize}_{(\mbox{w.r.t. }\br_+^K)} \quad \rank_{\mathrm{tc}}(\tensorX) \quad \text{\rm subject to} \quad \mc G[\tensorX] = \mc G[ \tensorXo ]. \label{eqn:vector_non_convex}
\end{eqnarray}
In vector optimization, a feasible point is called {\em Pareto optimal} if no other feasible point dominates it in every criterion. In a similar vein, we say that \eqref{eqn:vector_non_convex} recovers $\tensorXo$ if there does not exist any other tensor $\mcb{X}$ that is consistent with the observations and has no larger rank along each mode:


\begin{definition} \label{def: recover_nonconvex}
We call $\tensorXo$ recoverable by \eqref{eqn:vector_non_convex} if the set $$\{\tensorX' \neq \tensorXo \;|\; \mc{G}[\tensorX']=\mc{G}[\tensorXo], \; \rank_{\mathrm{tc}}(\tensorX')\preceq_{\br_+^K} \rank_{\mathrm{tc}}(\tensorXo)  \} = \emptyset.$$
\end{definition}
This is equivalent to saying that $\tensorXo$ is the unique optimal solution to the {\em scalar} optimization:
\begin{equation} \label{eqn:scalar_non_convex}
 \text{\rm minimize}_{\tensorX} \quad  \max_i \set{ \frac{\rank(\tensorX_{(i)})}{\rank(\tensorXo_{(i)})}} \qquad \mbox{\rm subject to} \quad \mc{G}[\tensorX]=\mc{G}[\tensorXo].
\end{equation}

The problems \eqref{eqn:vector_non_convex}-\eqref{eqn:scalar_non_convex} are not tractable. However, they do serve as a baseline for understanding how many generic measurements are required to recover $\mcb{X}_0$. The recovery performance of program \eqref{eqn:vector_non_convex} depends heavily on the properties of $\mc{G}$. Suppose \eqref{eqn:vector_non_convex} fails to recover $\tensorXo \in \mf{T}_r$. Then there exists another $\tensorX' \in \mf{T}_r$ such that $\mc{G}[\tensorX']=\mc{G}[{\tensorXo}]$. So, to guarantee that \eqref{eqn:vector_non_convex} recovers {\em any} $\tensorXo \in \mf{T}_r$, a necessary and sufficient condition is that $\mc{G}$ is injective on $\mf{T}_r$, which can be implied by the condition $\mathrm{null}(\mc G) \cap \mf T_{2r} = \set{\mb 0}$. Consequently, if $\mathrm{null}(\mc G) \cap \mf T_{2r} = \set{\mb 0}$, \eqref{eqn:vector_non_convex} will recover any $\mcb{X}_0 \in \mf T_r$. We expect this to occur when the number of measurements significantly exceeds the number of intrinsic degrees of freedom of a generic element of $\mf T_r$, which is $O( r^K + nrK )$. The following theorem shows that when $m$ is approximately twice this number, with probability one, $\mc G$ is injective on $\mf T_r$:

\begin{theorem}\label{thm:nonconvex} Whenever $m \ge (2r)^K + 2 nr K + 1$, with probability one, $\rm{null}(\mc{G})\cap \mf{T}_{2r}=\{\mb{0}\}$, and hence \eqref{eqn:vector_non_convex} recovers every $\tensorXo\in \mf{T}_r$.
\end{theorem}

The proof of Theorem \ref{thm:nonconvex} follows from a covering argument, which we establish in several steps. Let
\begin{equation}
\mf{S}_{2r} = \set{ \mcb{D} \mid \mcb{D}\in \mf{T}_{2r}, \norm{\mcb{D}}{F} = 1}.
\end{equation}
The following lemma shows that the required number of measurements can be bounded in terms of the exponent of the covering number for $\mf{S}_{2r}$, which can be considered as a proxy for dimensionality:
\begin{lemma}\label{lem:prob} Suppose that the covering number for $\mf S_{2r}$ with respect to Frobenius norm, satisfies
\begin{equation}
N(\mf S_{2r}, \norm{\cdot}{F}, \eps ) \;\le\; \left( \beta / \eps \right)^d,
\end{equation}
for some integer $d$ and scalar $\beta$ that does not depend on $\eps$. Then if $m \ge d+1$, with probability one $\nullspace{\mc G} \cap \mf S_{2r} = \emptyset$, which implies that $\nullspace{\mc{G}}\cap \mf{T}_{2r} = \{\mb 0\}$.
\end{lemma}

It just remains to find the covering number of $\mf S_{2r}$. We use the following lemma, which uses the triangle inequality to control the effect of perturbations in the factors of the Tucker decomposition
\begin{eqnarray}
[[\mcb C; \mb U_1, \mb U_2, \cdots, \mb U_K]]:= \mcb C \times_1 \mb U_1 \times_2 \mb U_2 \times_3 \cdots \times_K \mb U_K,
\end{eqnarray}
where the {\em mode-$i$ (matrix) product} of tensor $\mcb A$ with matrix $\mb B$ of compatible size, denoted as $\mcb A \times_i \mb B$, outputs a tensor $\mcb C$  such that $\mcb C_{(i)} = \mb B \mcb A_{(i)}$.

\begin{lemma} \label{lem:lipschitz} Let $\mcb{C}, \mcb{C}' \in \reals^{r_1, \dots, r_K}$, and $\mb U_1, \mb U_1' \in \reals^{n_1 \times r_1}, \dots, \mb U_K, \mb U_K' \in \reals^{n_K \times r_K}$ with $\mb U_i^* \mb U_i ={\mb U_i'}^* \mb U_i' =  \mb I$,  and $\norm{\mcb{C}}{F} = \norm{\mcb{C}'}{F} = 1$. Then
\begin{equation}
\norm{ [[ \mcb{ C }; \mb U_1, \dots, \mb U_K ]] - [[ \mcb{ C }'; \mb U'_1, \dots, \mb U'_K ]] }{F} \;\le\; \norm{\mcb{C} - \mcb{ C} '}{F} + \sum_{i=1}^K \norm{\mb U_i - \mb U_i'}.
\end{equation}
\end{lemma}
Using this result, we construct an $\eps$-net for $\mf S_{2r}$ by building $\eps / (K+1)$-nets for each of the $K+1$ factors $\mcb{C}$ and $\{\mb U_i\}$. The total size of the resulting $\eps$ net is thus bounded by the following lemma:

\begin{lemma} \label{lem:covering-new} $N(\mf S_{2r}, \norm{\cdot}{F}, \eps ) \;\le\; \left( 3 (K+1) / \eps \right)^{ (2r)^K + 2 n r K }$
\end{lemma}
With these observations in hand, Theorem \ref{thm:nonconvex} follows immediately.

\section{Convexification: Sum of Nuclear Norms?} \label{sec:SNN}
Since the nonconvex problem \eqref{eqn:vector_non_convex} is NP-hard for general $\mc{G}$, it is tempting to seek a convex surrogate. In matrix recovery problems, the nuclear norm is often an excellent convex surrogate for the rank \cite{fazel2002matrix, recht2010guaranteed,Gross2011-IT}. It seems natural, then, to replace the ranks in \eqref{eqn:vector_non_convex} with nuclear norms, and solve
\begin{eqnarray}
\text{minimize} \;\;  h(\tensorX):=\left(\norm{\tensorX_{(1)}}{*},\norm{\tensorX_{(2)}}{*},\cdots,\norm{\tensorX_{(K)}}{*}\right) \quad \mbox{subject to}\quad \mc{G}[\tensorX]=\mc{G}[\tensorXo].  \label{vector_convex}
\end{eqnarray}

Since $\norm{\tensorX_{(i)}}{*}$ is a convex function, the set $\mc H:= \bigcup_{\mcb{X} : \mc{G}[\tensorX] = \mc{G}[\tensorXo]} \set{\mcb{Y} \mid h(\mcb{Y}) \succeq h(\tensorX)}$ is convex. For any pareto optimal point $\widehat{\tensorX}$, there is a hyperplane supporting $\mc H$ passing through $h(\widehat{\tensorX})$, with normal vector $\mb\lambda\ge\mb0$. Therefore, $\widehat{\tensorX}$ is an optimal solution to the following scalar optimization:
\begin{eqnarray}
\text{\rm minimize} \quad \sum_{i=1}^K \lambda_i\|\tensorX_{(i)}\|_* \quad \text{\rm subject to} \quad \mc{G}[\tensorX] =\mc{G}[\tensorXo]. \label{eqn:scalar_convex}
\end{eqnarray}

The optimization \eqref{eqn:scalar_convex} was first introduced by \cite{liu2009tensor} and has been used successfully in applications in imaging \cite{Signoretto2011-SPL,Semerci2013-TIP,Kreimer2013-ICASSP,Li2010-ICIP,Ely2013-pp,Li2010-ECCV}. Similar convex relaxations have been considered in a number of theoretical and algorithmic works \cite{gandy2011tensor,signoretto2010nuclear,tomioka2010estimation,tomioka2011statistical,signoretto2013-ML}. It is not too surprising, then, that \eqref{eqn:scalar_convex} provably recovers the underlying tensor $\mcb{X}_0$, when the number of measurements $m$ is sufficiently large. For example, the following is a (simplified) corollary of results of Tomioka et.\ al.\ \cite{tomioka2010estimation}:\footnote{Tomioka et.\ al.\ also show noise stability when $m = \Omega( r n^{K-1} )$ and give extensions to the case where the $\ranktc{\mcb{X}_0} = (r_1, \dots, r_K)$ differs from mode to mode.}
\begin{corollary}[of \cite{tomioka2010estimation}, Theorem 3] \label{cor:SNN} Suppose that $\mcb{X}_0$ has Tucker rank $(r,\dots, r)$, and  $m \ge C r n^{K-1}$. With high probability, $\mcb{X}_0$ is an optimal solution to \eqref{eqn:scalar_convex}, with each $\lambda_i = 1$. Here, $C$ is numerical.
\end{corollary}

This result shows that there {\em is} a range in which \eqref{eqn:scalar_convex} succeeds: loosely, when we undersample by at most a factor of $m/N \sim r/n$. However, the number of observations $m \sim r n^{K-1}$ is significantly larger than the number of degrees of freedom in $\mcb{X}_0$, which is on the order of $r^K + n r K$. Is it possible to prove a better bound for this model? Unfortunately, we show that in general $O(rn^{K-1})$ measurements are also {\em necessary} for reliable recovery using \eqref{eqn:scalar_convex}:

\newcommand{\tensors}[1][]{\mf T_{#1}}

\begin{theorem} \label{thm:SNNM_main} Let $\mcb{X}_0 \in \tensors[r]$ be nonzero. Set $\kappa = \min_i \set{ \norm{(\mcb{X}_0)_{(i)}}{*}^2 / \norm{\mcb{X}_0}{F}^2 } \times n^{K-1}$. Then if the number of measurements $m \le \kappa - 2$,  $\tensorXo$ is not the unique solution to \eqref{eqn:scalar_convex}, with probability at least $1- 4\exp (-\frac{(\kappa - m -2)^2}{16 ( \kappa - 2)})$. Moreover, there exists $\mcb{X}_0 \in \tensors[r]$ for which $\kappa = r n^{K-1}$.
\end{theorem}

This implies that Corollary \ref{cor:SNN} (and other results of \cite{tomioka2010estimation}) is essentially tight. Unfortunately, it has negative implications for the efficacy of the sum of nuclear norms in \eqref{eqn:scalar_convex}: although a generic element $\mcb{X}_0$ of $\tensors[r]$ can be described using at most $r^K + nr K$ real numbers, we require $\Omega (r n^{K-1})$ observations to recover it using \eqref{eqn:scalar_convex}. Theorem \ref{thm:SNNM_main} is a direct consequence of a much more general principle underlying multi-structured recovery, which is elaborated next.

\subsection*{Recovering objects with multiple structures}

The poor behavior of \eqref{eqn:scalar_convex} is actually an instance of a much more general phenomenon, first discovered by Oymak et.\ al.\ \cite{oymak2012simultaneously}. Our target tensor $\mcb{X}_0$ has {\em multiple} low-dimensional structures simultaneously: it is low-rank along {\em each} of the $K$ modes. In practical applications, many other such {\em simultaneously structured} objects may be of interest -- for example, matrices that are simultaneously sparse and low-rank \cite{richard2012estimation, oymak2012simultaneously}. To recover such a simultaneously structured object, it is tempting to build a convex relaxation by combining the convex relaxations for each of the individual structures. In the tensor case, this yields \eqref{eqn:scalar_convex}. Surprisingly, this combination is often not significantly more powerful than the best single regularizer \cite{oymak2012simultaneously}. We obtain Theorem \ref{thm:SNNM_main} as a consquence of a new, general result of this nature, using a geometric framework introduced in \cite{amelunxen2013living}. Compared to the proof strategy in \cite{oymak2012simultaneously}, this approach has a clearer geometric intuition, covers a more general class of regularizers and yields sharper bounds.

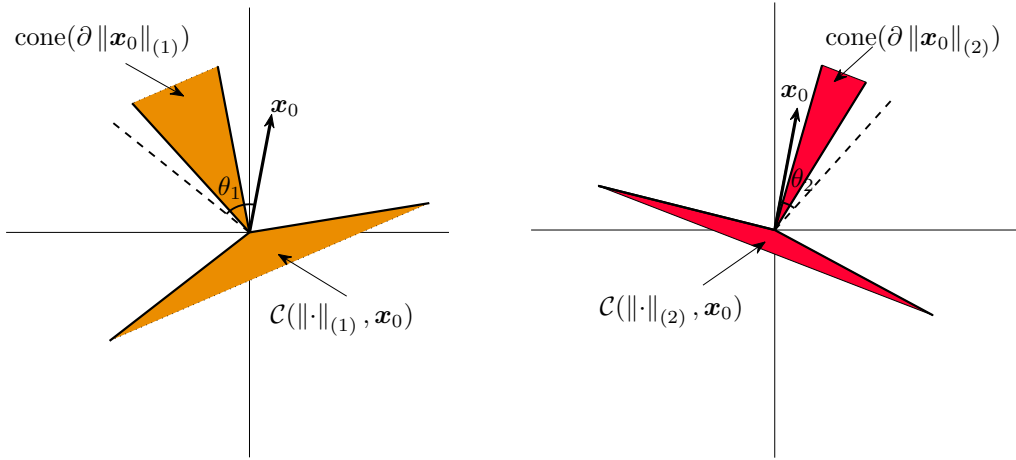
\begin{figure}[h]
\setlength{\abovecaptionskip}{0pt}
\setlength{\belowcaptionskip}{0pt} \centerline{
\begin{minipage}{3.5in}
\centerline{
\begin{overpic}[width=3in,unit=1mm]{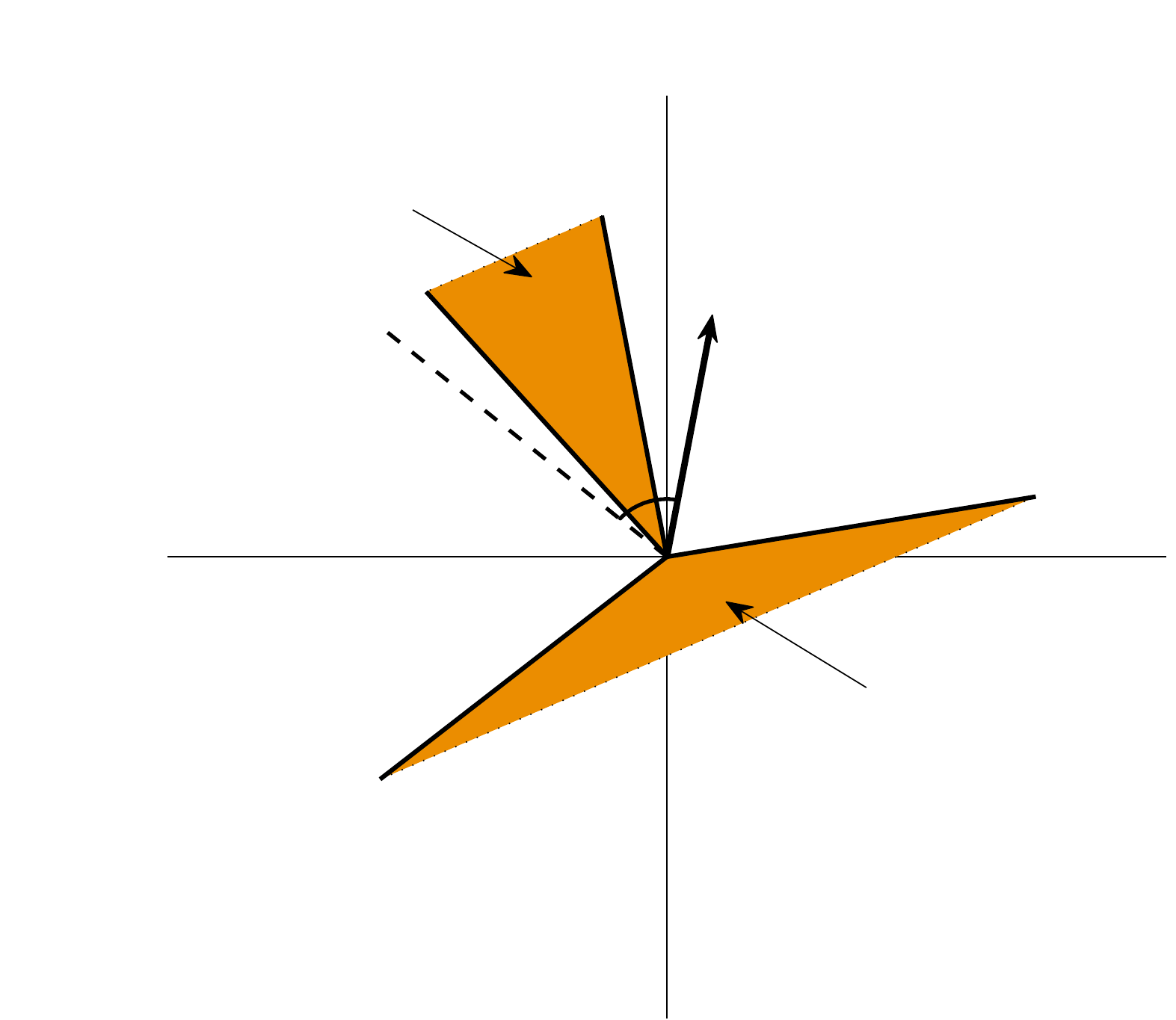}
    \put(12,55){$\text{cone}(\partial \norm{\mb x_0}{(1)})$}
    \put(46,46){$\mb x_0$}
    \put(39,35){$\theta_1$}
    \put(46,18){$\mc C(\norm{\cdot}{(1)}, \mb x_0 )$}
\end{overpic}
}
\end{minipage}
\hspace{-15mm}
\begin{minipage}{3.5in}
\begin{overpic}[width=3in,unit=1mm]{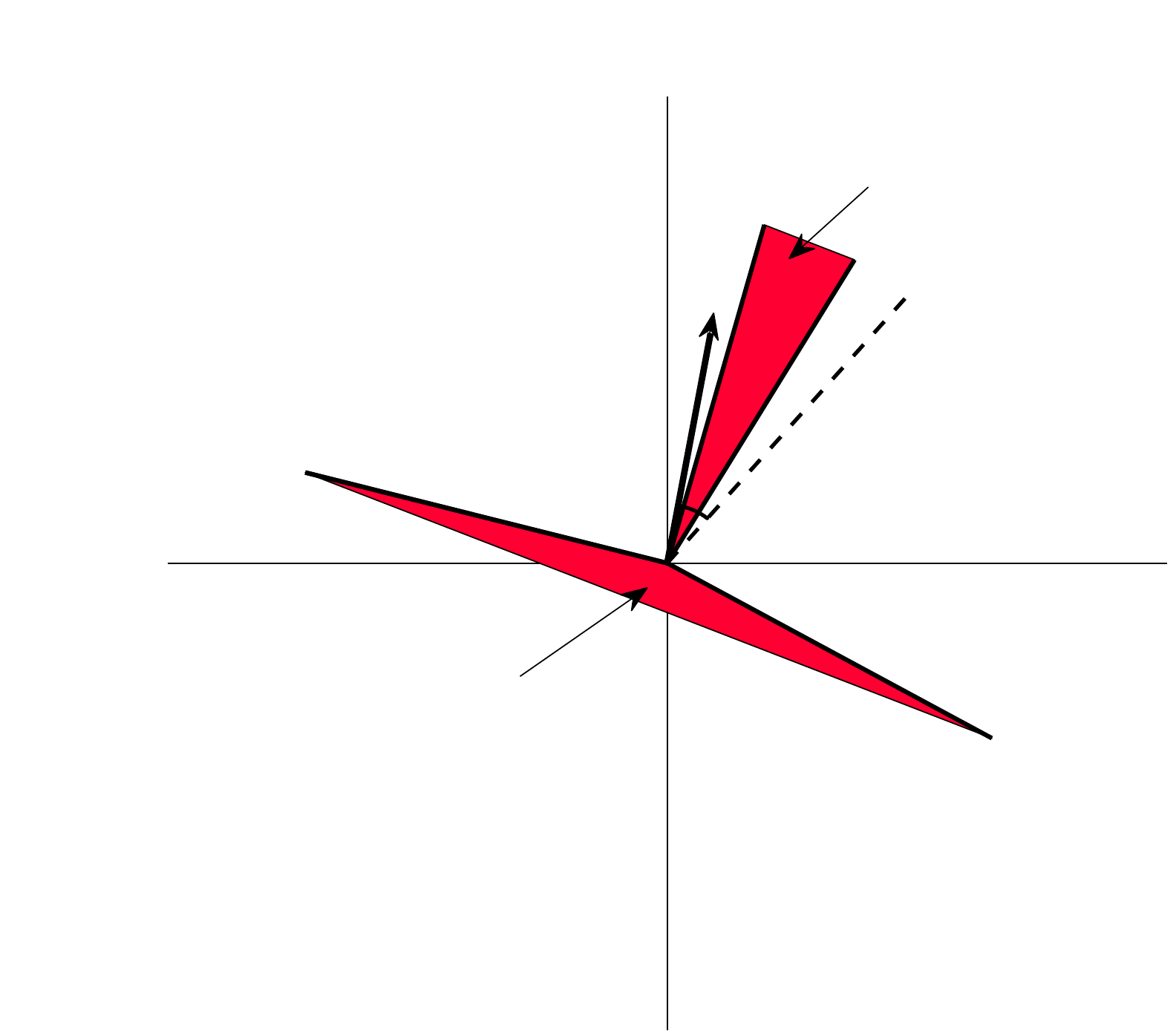}
    \put(50,55){$\text{cone}(\partial \norm{\mb x_0}{(2)})$}
    \put(44,48){$\mb x_0$}
    \put(45.4,36){$\theta_2$}
    \put(20,19){$\mc C(\norm{\cdot}{(2)}, \mb x_0 )$}
\end{overpic}
\end{minipage}
}
\caption{{\bf Cones and their polars for convex regularizers $\norm{\cdot}{(1)}$ and $\norm{\cdot}{(2)}$ respectively.} Suppose our $\mb x_0$
has two sparse structures simultaneously. Regularizer $\norm{\cdot}{(1)}$ has a larger conic hull of subdifferential at $\mb x_0$, i.e. $\mbox{cone}(\partial \norm{\mb x_0}{(1)})$, which results in a smaller descent cone. Thus minimizing $\norm{\cdot}{(1)}$ is more likely to recover $\mb x_0$ than minimizing $\norm{\cdot}{(2)}$. Consider convex regularizer $f(\mb x) = \norm{\mb x_0}{(1)} + \norm{\mb x_0}{(2)}$.
Suppose as depicted, $\theta_1\ge \theta_2$. Then both $\mbox{cone}(\partial \norm{\mb x_0}{(1)})$ and $\mbox{cone}(\partial \norm{\mb x_0}{(2)})$ are in the circular cone $\mbox{circ}(\mb x_0, \theta_1)$. Thus we have:
 $\mbox{cone}\big{(}\partial f(\mb x_0)\big{)} = \mbox{cone}(\partial \norm{\mb x_0}{(1)}+\partial \norm{\mb x_0}{(2)}) \subseteq \mbox{conv}\big{\{}\mbox{circ}(\mb x_0, \theta_1), \mbox{circ}(\mb x_0, \theta_2) \big{\}} = \mbox{circ}(\mb x_0, \theta_1)$.}
\label{fig:cone_polar}
\end{figure}

Consider a signal  $\mb x_0\in \reals^n$  having $K$ low-dimensional structures simultaneously (e.g. sparsity, low-rank, etc.)\footnote{$\mb x_0$ is the underlying signal of our interest (perhaps after vectorization).}.
Let $\norm{\cdot}{(i)}$ be the penalty norms corresponding to the $i$-th structure (e.g. $\ell_1$, nuclear norm).
 Consider the composite norm optimization
\begin{equation}\label{eqn:main_prob}
    \min_{\mb x\in \reals^n} f(x):= \lambda_1 \norm{\mb x}{(1)}+\lambda_2 \norm{\mb x}{(2)}+\cdots+\lambda_K \norm{\mb x}{(K)} \qquad \mbox{subject to} \quad \mc{G}[\mb x]=\mc{G}[\mb x_0],
\end{equation}
where $\mc{G}[\cdot]$ is a Gaussian measurement operator, and $\mb \lambda > \mb 0$. Is $\mb x_0$ the unique optimal solution to \eqref{eqn:main_prob}? Recall that the descent cone of a function $f$ at a point $\mb x_0$ is defined as
\begin{equation}
\mc C (f, \mb x_0) = \mathrm{cone}\set{ \mb v \mid f( \mb x_0 + \mb v ) \le f( \mb x_0 ) },
\end{equation}
which, in short, will be denoted as $\mc C$. Then $\mb x_0$ is the unique optimal solution if and only if $\mathrm{null}(\mc G) \cap \mc C = \set{ \mb 0 }$. Conversely, recovery fails if $\mathrm{null}(\mc G)$ has nontrivial intersection with $\mc C$. If $\mc G$ is a Gaussian operator, $\mathrm{null}(\mc G)$ is a uniformly oriented random subspace of dimension $n - m$. This random subspace is more likely to have nontrivial intersection with $\mc C$ if $\mc C$ is ``large,'' in a sense we will make precise.  The polar of $\mc C$ is $\mc C^o = \mathrm{cone}\big{(} \partial f(\mb x_0) \big{)}$. Because polarity reverses inclusion, we expect that $\mc C$ will be ``large'' whenever $\mc C^o$ is ``small''. Figure \ref{fig:cone_polar} visualizes this geometry.

To control the size of $C^o$, first consider a single norm $\norm{\cdot}{\diamond}$, with dual norm $\norm{\cdot}{\diamond}^*$. Suppose that $\norm{\cdot}{\diamond}$ is $L$-Lipschitz: $\norm{\mb x}{\diamond} \le L \norm{\mb x}{2}$ for all $\mb x$. Then $\norm{\mb x}{2} \le L \norm{\mb x}{\diamond}^*$ for all $\mb x$ as well. Noting that
$$\partial \norm{\cdot}{\diamond}(\mb x) = \set{ \mb v \mid \<\mb v, \mb x \> = \norm{\mb x}{\diamond}, \; \norm{\mb v}{\diamond}^* \le 1 },$$
for any $\mb v \in \partial \norm{\cdot}{\diamond}(\mb x_0)$, we have
\begin{equation}
\frac{\< \mb v, \mb x_0\>}{ \norm{\mb v}{2} \norm{\mb x_0}{2} } \;\ge\;  \frac{\norm{\mb x_0}{\diamond}}{L \norm{\mb v}{\diamond}^* \norm{\mb x_0}{2}}  \;\ge\; \frac{\norm{\mb x_0}{\diamond}}{L \norm{\mb x_0}{2}}.
\end{equation}
A more geometric way of summarizing this is as follows: for $\mb x \ne \mb 0$, let
\begin{equation}
\mathrm{circ}(\mb x, \theta) = \set{ \mb z \mid \angle(\mb z, \mb x) \le \theta },
\end{equation}
and denote the {\em circular cone} with axis $\mb x$ and angle $\theta$. Then if $\mb x_0 \ne \mb 0$, and $\theta = \cos^{-1} ( \norm{\mb x_0}{\diamond} / L \norm{\mb x_0}{2} )$,
\begin{equation}
\partial \norm{\cdot}{\diamond}(\mb x_0) \subseteq \mathrm{circ}\left(\mb x_0, \theta \right).
\end{equation}
Table \ref{tab:angles} describes the angle parameters $\theta$ for various structure inducing norms. Notice that in general, more complicated $\mb x_0$ leads to smaller angles $\theta$. For example, if $\mb x_0$ is a $k$-sparse vectors with entries all of the same magnitude, and $\norm{\cdot}{\diamond}$ the $\ell^1$ norm, $\cos^2 \theta = k/n$. As $\mb x_0$ becomes more dense, $\partial \norm{\cdot}{\diamond}$ is contained in smaller and smaller circular cones.

For $f = \sum_i \lambda_i \norm{\cdot}{(i)}$, notice that every element of $\partial f(\mb x_0)$ is a conic combination of elements of the $\partial \norm{\cdot}{(i)}(\mb x_0)$. Since each of the $\partial \norm{\cdot}{(i)}(\mb x_0)$ is contained in a circular cone with axis $\mb x_0$,  $\partial f(\mb x_0)$ is also contained in a circular cone:
\begin{lemma}\label{lem:circ} Suppose that $\norm{\cdot}{(i)}$ is $L_i$-Lipschitz. For $\mb x_0 \ne \mb 0$, set $\theta_i = \cos^{-1}\left( \norm{\mb x_0}{(i)} / L \norm{\mb x_0}{2} \right)$. Then
\begin{equation}
\partial f( \mb x_0 ) \subseteq \mathrm{circ}\left( \mb x_0 , \max_{i=1 \dots K} \theta_i \right).
\end{equation}
\end{lemma}
\noindent So, the subdifferential of our combined regularizer $f$ is contained in a circular cone whose angle is given by the largest of the $\theta_i$.

\begin{table}
\centerline{
\begin{tabular}{|l|c|c|c|c|c|}
\hline
{\bf Object} & {\bf Complexity Measure} & {\bf Relaxation} & $\cos^2 \theta$ & $\kappa$ \\
\hline \hline
Sparse $\mb x \in \reals^n$ & $k = \norm{\mb x}{0}$ & $\norm{\mb x}{1}$ & $[\tfrac{1}{n},\tfrac{ k }{ n }]$ & $[1,k]$ \\
\hline
Column-sparse $\mb X \in \reals^{n_1 \times n_2}$ & $c = \# \set{ j \mid \mb X \mb e_j \ne \mb 0 }$ & $\sum_{j} \norm{\mb X \mb e_j}{2}$ & $[ \tfrac{1}{n_2} , \tfrac{c}{n_2} ]$ & $[ n_1 , c n_1 ]$ \\
\hline
Low-rank  $\mb X \in \reals^{n_1 \times n_2}$ {\scriptsize ($n_1 \ge n_2$)} & $r = \mathrm{rank}(\mb X)$ & $\norm{\mb X}{*}$ & $[ \tfrac{1}{n_2}, \tfrac{r}{n_2}]$ & $[ n_1, r n_1 ]$ \\
\hline
\hline
{Low-rank $\mcb{X} \in \reals^{n \times n \times \dots \times n}$} & $\ranktc{ \mcb{X} }$ & $\sum_i \norm{\mcb{X}_{(i)} }{*}$ & $ [\tfrac{1}{n},\tfrac{r}{n}] $ & $[ n^{K-1}, r n^{K-1}]$ \\
\hline
 & $\ranktc{\mcb{X}}$ & $\norm{\mcb{X}}{\square}$ & $[  (\tfrac{1}{n})^{\lfloor \tfrac{K}{2} \rfloor}, (\tfrac{r}{n})^{\lfloor \tfrac{K}{2} \rfloor} ]$ & $[ n^{\lceil \tfrac{K}{2} \rceil}, r^{\lfloor \tfrac{K}{2} \rfloor} n^{\lceil\tfrac{K}{2}\rceil}]$  \\
\hline
\end{tabular}}
\caption{{\bf Concise models and their surrogates.} For each norm $\norm{\cdot}{\diamond}$, the third column describes the range of achievable angles $\theta$. Larger $\cos \theta$ corresponds to a smaller $C^o$, a larger $C$, and hence a larger number of measurements required for reliable recovery. The fourth line is the sum of nuclear norms; the last line is the square norm introduced in Section \ref{sec:square}.} \label{tab:angles}
\end{table}

How does this behavior affect the recoverability of $\mb x_0$ via \eqref{eqn:main_prob}? The informal reasoning above suggests that as $\theta$ becomes smaller, the descent cone $\mc C$ becomes larger, and we require more measurements to recover $\mb x_0$. This can be made precise using an elegant framework introduced by Amelunxen et.\ al.\ \cite{amelunxen2013living}. They define the {\em statistical dimension} of the convex cone $\mc C$ to be the expected norm of the projection of a standard Gaussian vector onto $C$:
\begin{equation}
\delta(\mc C) \doteq \expect[\mb g \sim_{\mathrm{i.i.d.}} \mc N(0,1)]{\norm{\mc P_{\mc C} (\mb g)}{2}^2}.
\end{equation}
Using tools from spherical integral geometry, \cite{amelunxen2013living} shows that for linear inverse problems with Gaussian measurements, a sharp phase transition in recoverability occurs around $m = \delta(\mc C)$. We will need only one side of their result; for more details see \cite{amelunxen2013living}. We state a slight variant here:
\begin{corollary}\label{cor:kinematic} Let $\mc G : \reals^n \to \reals^m$ be a Gaussian operator, and $\mc C$ a convex cone. Then if $m \le \delta(\mc C)$,
\begin{equation}
\prob{\mc C \cap \mathrm{null}(\mc G) = \set{ \mb 0 } } \;\le\; 4 \exp\left( - \frac{ (\delta(\mc C) - m)^2}{16 \delta(\mc C) } \right).
\end{equation}
\end{corollary}
To apply this result to our problem, we lower bound the statistical dimension  $\delta(\mc C)$, of the descent cone $\mc C$ of $f$ at $\mb x_0$. Using the Pythagorean theorem, monotonicity of $\delta(\cdot)$, and Lemma \ref{lem:circ}, we calculate
\begin{equation}\label{eqn:polar_cone}
\delta(\mc C) \;=\; n - \delta(\mc C^o) \;=\; n - \delta\left( \mathrm{cone}( \partial f (\mb x_0) ) \right) \;\ge\; n - \delta( \mathrm{circ}(\mb x_0, \max_i \theta_i ) ).
\end{equation}
Moreover, using the properties of statistical dimension, we are able to prove an upper bound for the statistical dimension of circular cone, which improves the constant in existing results \cite{amelunxen2013living, mccoy2013geometric}.
\begin{lemma}\label{lem:bd_cc}
$\delta( \mathrm{circ}(\mb x_0, \theta) ) \le n \sin^2 \theta + 2$.
\end{lemma}
Finally, by combining \eqref{eqn:polar_cone} and Lemma \ref{lem:bd_cc}, we have
 $\delta(C) \ge n \min_i \cos^2 \theta_i - 2$. Using Corollary \ref{cor:kinematic}, we obtain:
\begin{theorem} \label{thm:sim-struc}
Let $\mb x_0 \ne \mb 0$. Suppose that for each $i$, $\norm{\cdot}{(i)}$ is $L_i$-Lipschitz. Set $$\kappa_i \;=\; \frac{n \norm{\mb x_0}{(i)}^2}{L_i^2 \norm{\mb x_0}{2}^2} \;=\; n \cos^2( \theta_i ),$$
and $\kappa = \min_i \kappa_i$. Then if $m \le \kappa - 2$,
\begin{equation}
\prob{ \mb x_0 \; \text{\rm is the unique optimal solution to \eqref{eqn:main_prob}} } \le 4 \exp\left( - \frac{( \kappa - m - 2)^2 }{16\, ( \kappa - 2 )} \right).
\end{equation}
\end{theorem}

Thus, for reliable recovery, the number of measurements needs to be at least proportional to $\kappa$.\footnote{E.g., if $m = (\kappa - 2)/2$, the probability of success is at most $4 \exp( - (\kappa - 2) / 64 )$.} Notice that $\kappa = \min_i \kappa_i$ is determined by only the best of the structures. Per Table \ref{tab:angles}, $\kappa_i$ is often on the order of the number of degrees of freedom in a generic object of the $i$-th structure. For example, for a $k$-sparse vector whose nonzeros are all of the same magnitude, $\kappa = k$.

Theorem \ref{thm:sim-struc} together with Table \ref{tab:angles} leads us to the phenomenon that recently discovered by Oymak et.\ al.\ \cite{oymak2012simultaneously}: for recovering objects with multiple structures, a combination of structure-inducing norms tends to be not significantly more powerful than the best individual structure-inducing norm. As we demonstrate, this general behavior follows a clear geometric interpretation that the subdifferential of a norm at $\mb x_0$ is contained in a relatively small circular cone with central axis $\mb x_0$.

We can specialize Theorem \ref{thm:sim-struc} to low-rank tensors as follows: if $\mcb{X}$ is a $K$-mode $n \times n \times \dots \times n$ tensor of Tucker rank $(r, r, \dots, r)$, then for each $i$, $\norm{\mcb{X}}{(i)} \doteq \norm{ \mcb{X}_{(i)} }{*}$ is $L= \sqrt{n}$-Lipschitz. Hence, \begin{equation}
\kappa = \min_i \set{\norm{\mcb{X}_{(i)}}{*}^2 / \norm{\mcb{X}}{F}^2  } \, n^{K-1}.
\end{equation}
The term in brackets lies between $1$ and $r$, inclusive. For example, if $\mcb X = [[ \mcb{C}, \mb U_1, \dots, \mb U_K ]]$, with $\mb U_i^T \mb U_i = \mb I$ and $\mcb{C}$ supersymmetric ($\mcb{C}_{i_1 \dots i_K} = \indicator{i_1 = i_2 = \dots = i_K}$), then this term is equal to $r$.

\section{A Better Convexification: Square Norm} \label{sec:square}
The number of measurements promised by Corollary \ref{cor:SNN} and Theorem \ref{thm:SNNM_main} is actually the same (up to constants) as the number of measurements required to recover a tensor $\mcb{X}_0$ which is low-rank along just one mode. Since matrix nuclear norm minimization correctly recovers a $n_1 \times n_2$  matrix of rank $r$ when $m \ge C r (n_1 + n_2)$ \cite{chandrasekaran2012convex}, solving
\begin{equation}
\text{minimize} \; \|\tensorX_{(1)}\|_* \quad \text{subject to} \quad \mc{G}[\tensorX] = \mathcal{G}[\tensorX_0]
\end{equation}
also exactly recovers $\mcb{X}_0$ with high probability when $m \ge C r n^{K-1}$.

This suggests a more mundane explanation for the difficulty with \eqref{eqn:scalar_convex}: the term $r n^{K-1}$ comes from the need to reconstruct the right singular vectors of the $n \times n^{K-1}$ matrix $\mcb{X}_{(1)}$. If we had some way of matricizing a tensor that {\em produced a more balanced (square) matrix} and also {\em preserved the low-rank property}, we could substantially reduce this effect, and reduce the overall sampling requirement. In fact, this is possible when the order $K$ of $\mcb{X}_0$ is four or larger.

\newcommand{\rankcp}[1]{\mathrm{rank}_{\mathrm{cp}}\left( #1 \right)}

For $\mb A\in \reals^{m_1 \times n_1}$, and integers $m_2$ and $n_2$ satisfying $m_1n_1=m_2n_2$, the reshaping operator $\mbox{reshape}(\mb A, m_2, n_2)$ returns a $m_2\times n_2$ matrix whose elements are taken columnwise from $\mb A$. This operator rearranges elements in $\mb A$ and leads to a matrix of different shape. In the following, we reshape matrix $\tensorX_{(1)}$ to a more square matrix while preserving the low-rank property. Let $\tensorX \in \reals^{n_1\times n_2\times \cdots \times n_K}$. Select $j\in [K]:=\{1,\; 2,\; \cdots, \; K\}$. Then we define matrix $\mcb{X}_{[j]}$ as

$$\mcb{X}_{[j]}=\mbox{reshape}\Bigl(\mcb{X}_{(1)},\;\prod_{i=1}^j  n_i,\;\prod_{i=j+1}^K n_i\Bigr).$$

We can view $\mcb{X}_{[j]}$ as a natural generalization of the standard tensor matricization. When $j=1$, $\mcb{X}_{[j]}$ is nothing but $\mcb{X}_{(1)}$. However, when some $j > 1$ is selected, $\mcb{X}_{[j]}$ becomes a more balanced matrix. This reshaping also preserves some of the algebraic structures of $\mcb{X}$. In particular, we will see that if $\mcb{X}$ is a low-rank tensor (in either the CP or Tucker sense), $\mcb{X}_{[j]}$ will be a low-rank matrix.

\begin{lemma} \label{lem: arrangement}
(1) If $\mcb{X}$ has CP decomposition $\mcb{X}=\sum_{i=1}^r \lambda_i \mb a_{i}^{(1)}\circ \mb a_{i}^{(2)}\circ \cdots \circ \mb a_{i}^{(K)}$, then
\begin{equation}\label{eqn:CP_rearrange}
\mcb{X}_{[j]} = \sum_{i=1}^r \lambda_i(\mb a_i^{(j)}\otimes \mb a_i^{(j-1)}\otimes \cdots \otimes \mb a_i^{(1)})\circ  (\mb a_{i}^{(K)} \otimes \mb a_{i}^{(K-1)} \cdots \otimes \mb a_{i}^{(j+1)}).
\end{equation}
(2) If $\mcb{X}$ has Tucker decomposition
$\mcb{X}=\mcb{C}\times_1 \mb U_1 \times_2 \mb U_2 \times_3 \cdots \times_K \mb U_K$, then
\begin{equation}
\mcb{X}_{[j]} = (\mb U_j \otimes \mb U_{j-1} \otimes \cdots \otimes \mb U_1) \,\mcb{C}_{[j]}\, (\mb U_K \otimes \mb U_{K-1} \otimes \cdots \otimes \mb U_{j+1})^*.
\end{equation}
\end{lemma}

Using Lemma \ref{lem: arrangement} and the fact that $\rank( \mb A \tensor \mb B ) = \rank(\mb A) \, \rank(\mb B)$, we obtain:

\begin{lemma}\label{lem:ranks}
Let $\ranktc{\mcb{X}} = (r_1,r_2,\cdots,r_K)$, and $\rankcp{\mcb{X}} = r_{\mathrm{cp}}$. Then $\mathrm{rank}( \mcb{X}_{[j]}) \le r_{\mathrm{cp}}$, and $\mathrm{rank}( \mcb{X}_{[j]}) \le \min\Bigl\{ \; \prod_{i=1}^j r_i, \, \prod_{i=j+1}^K r_i \; \Bigr\}$.
\end{lemma}

Thus, $\tensorX_{[j]}$ is not only more balanced but also maintains the low-rank property of tensor $\tensorX$. In the following, we show how this new matricization can lead to better relaxations for tensor recovery. For ease of discussion, we assume $\tensorX$ has the same length (say $n$) along each mode and has Tucker rank $(r,r,\cdots,r)$.  We write $\tensorX_\square \doteq \tensorX_{[\lfloor \frac{K}{2} \rfloor]}$  and call $\norm{\mcb{X}}{\square} := \norm{\tensorX_\square}{\star}$ the {\em square norm} of tensor $\mcb X$. Since $\tensorX_\square$ is low-rank, we can attempt to recover $\mcb{X}$ by solving
\begin{eqnarray}
\label{eqn:square_norm}
\text{\mbox minimize} \quad \|\tensorX \|_\square \label{Improvement_convex} \quad \text{\mbox subject to} \quad \mc{G}[\tensorX]=\mc{G}[\tensorXo].
\end{eqnarray}
Using Lemma \ref{lem:ranks} and Proposition 3.11 of \cite{chandrasekaran2012convex}, we can prove that this relaxation exactly recovers $\mcb{X}_0$, when the number of measurements is sufficienly large:

\begin{theorem} \label{thm:square-recovery}
(1) If $\tensorXo$ has CP rank $r$, using \eqref{eqn:square_norm}, $m \ge Crn^{\lceil \frac{K}{2} \rceil}$ is sufficient to recover $\tensorXo$ with high probability.
(2) If $\tensorXo$ has Tucker rank $(r,r,\cdots,r)$, using \eqref{eqn:square_norm}, $m \ge Cr^{\lfloor \frac{K}{2} \rfloor}n^{\lceil \frac{K}{2} \rceil}$ is sufficient to recover $\tensorXo$ with high probability.
\end{theorem}

Compared with $\Omega(rn^{K-1})$ measurements required by the sum-of-nuclear-norms model, the sample complexity, $O(r^{\lfloor \frac{K}{2} \rfloor}n^{\lceil \frac{K}{2} \rceil})$, required by the square reshaping \eqref{eqn:square_norm}, is always within a constant of it, much better for small $r$ and $K\ge 4$ -- e.g., by a multiplicative factor of $n^{\lfloor K/ 2 \rfloor - 1}$ when $r$ is a constant. This is a significant improvement. However, there are also two clear limitations. First, no improvement is obtained for the case $K=3$. Second, the improved sample complexity in Theorem \ref{thm:square-recovery} is still suboptimal compared to the nonconvex model \eqref{eqn:vector_non_convex}.

It is also worth noting that for tensors with different lengths or ranks, Theorem \ref{thm:SNNM_main} and Theorem \ref{thm:square-recovery} can be easily modified. It remains true that for a large class of tensors, our square reshaping is capable of reducing the number generic measurements required by SNN model. However, the comparison between sum-of-nuclear-norms and square norm becomes quite subtle then. Concrete instances can be definitely constructed so that square norm model does not have any advantage over the SNN model even for $K>3$ (e.g. a tensor of size $1000 \times 10 \times 10 \times 10$ with Tucker rank $(1,1,1,1)$). On the other hand, our square norm model can sometimes be blessed by unbalanced tensors.  For example, consider a tensor of size $4 \times 9 \times 12 \times 3$ with Tucker rank $(2,2,1,1)$. Then our reshaping matrix is a $36 \times 36$ square matrix with rank $1$, which is a matrix with very good (perfect) conditions.


\section{Simulation Results for Tensor Completion} \label{sec:simulation}
Tensor completion attempts to reconstruct the low-rank tensor $\mcb X_0$ based on observations over a subset of its entries $\Omega$. By imposing appropriate incoherence conditions (and modifying slightly arguments in \cite{Gross2011-IT}), it is possible to prove recovery guarantees for each of the following programs:
\begin{eqnarray}
\mbox{minimize} \quad \sum_{i=1}^K \lambda_i\|\tensorX_{(i)}\|_* \quad \mbox{subject to} \quad \mc P_\Omega[\mcb X]=\mc P_\Omega[\mcb X_0];\label{eqn:SSN_completion}\\
\mbox{minimize} \quad \norm{\mcb X}{\square} \quad \mbox{subject to} \quad \mc P_\Omega[\mcb X]=\mc P_\Omega[\mcb X_0].\label{eqn:square_completion}
\end{eqnarray}
Unlike the recovery problem under Gaussian random measurements, due to the lack of sharp upper bounds, we have no proof that our square norm formulation outperforms the SNN model here. However, our simulation results below strongly suggest that \eqref{eqn:square_completion} also performs much better than \eqref{eqn:SSN_completion} for tensor completion case.

\begin{figure}[h]
\centerline{
\begin{minipage}{3.5in}
\centerline{\includegraphics[width=2.75in]{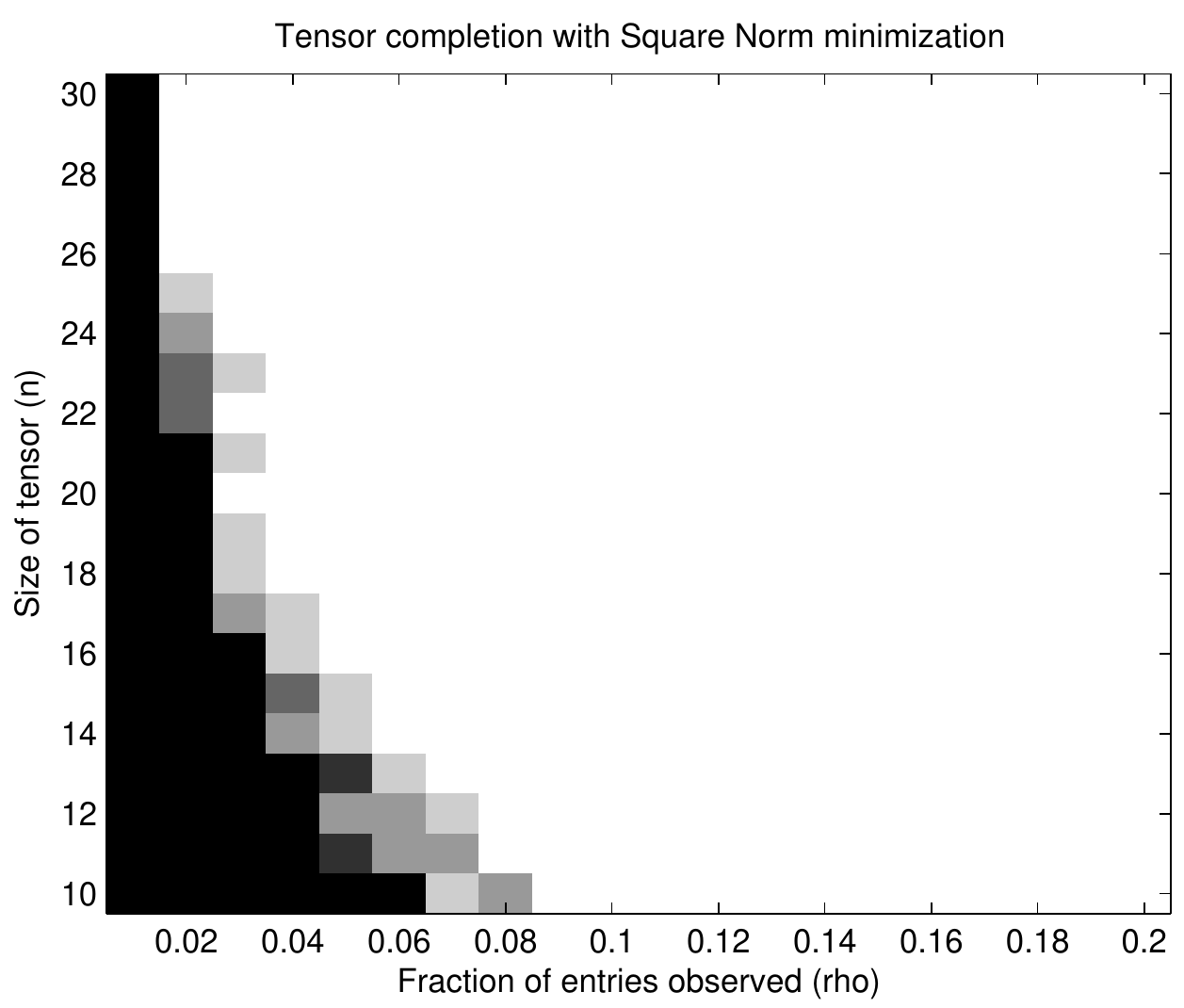}}
\end{minipage}
\hspace{-5mm}
\begin{minipage}{3.5in}
\centerline{\includegraphics[width=2.75in]{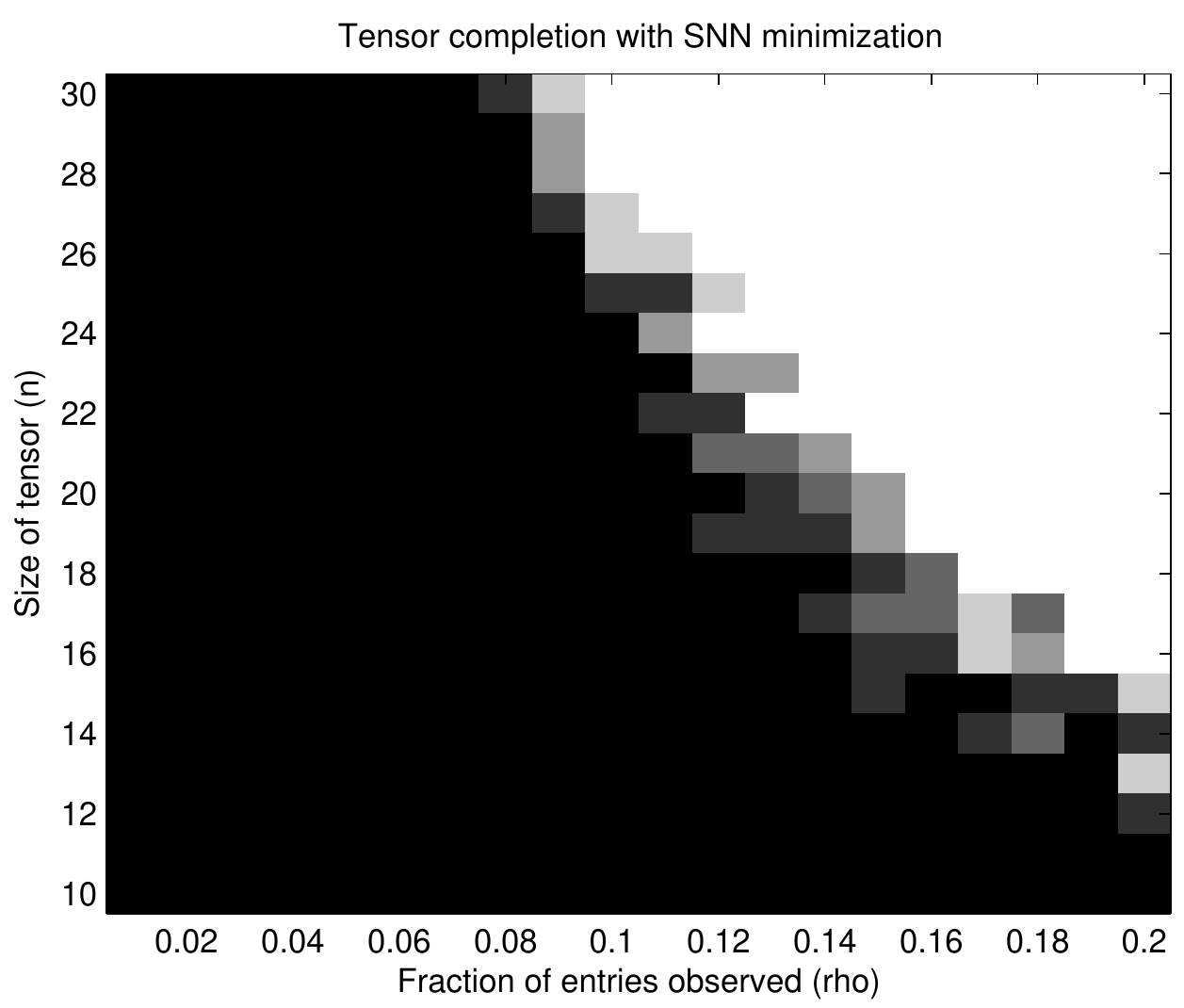} }
\end{minipage}
}
\caption{{\textbf{Tensor completion.}} The colormap indicates the fraction of correct recovery, which increases with brightness from certain failure (black) to certian success (white).}
\label{fig:tensor_completion}
\end{figure}

Our experiment is set up as follows. We generate a $4$-way tensor, $\mcb X_0 \in \reals^{n\times n\times n\times n}$, as $$\mcb X_0 = [[\mcb C_0; \mb U_1,  \mb U_2,  \mb U_3,  \mb U_4]] = {\mcb C_0} \times_1 \mb U_1 \times_2 \mb U_2 \times_3 \mb U_3 \times_4  \mb U_4,$$
where the core tensor $\mcb C_0\in \reals^{1\times 1\times 2\times 2}$ has i.i.d. standard Gaussian entries, and matrices $\mb U_1$, $\mb U_2 \in \reals^{n\times 1}$ and matrices $\mb U_3$, $\mb U_4 \in \reals^{n\times 2}$, satisfying $\mb U_i^* \mb U_i = \mb {I}$, are drawn uniformly at random (by the command $\mbox{orth}(\mbox{randn}(\cdot,\cdot))$ in Matlab). The observed entries are chosen uniformly with ratio $\rho$. We increase the problem size $n$ from $10$ to $30$ with increment $1$, and the observation ratio $\rho$ from $0.01$ to $0.2$ with increment $0.01$. For each $(\rho, n)$-pair, we simulate $5$ test instances and declare a trial to be successful if the recovered $\mcb X^\star$ satisfies $\frac{\norm{\mcb X^\star-\mcb X_0}{F}}{\norm{\mcb X_0}{F}}\le 10^{-2}$. The optimization problems are solved using efficient first-order methods. Since \eqref{eqn:square_completion} is in the form of standard matrix completion, we use the Augmented Lagrangian Method (ALM) proposed in \cite{lin2010augmented} to solve it. For the sum of nuclear norms minimization \eqref{eqn:SSN_completion} with $\lambda_i=1$, we implement the accelerated linearized Bregman algorithm \cite{huang2013accelerated}, of which we include a detailed discussion  in the appendix. Figure \ref{fig:tensor_completion} plots the fraction of correct recovery for each pair (black $=0\%$ and white $=100\%$). Clearly much larger white region is produced by square norm, which empirically suggests that \eqref{eqn:square_completion} outperforms \eqref{eqn:SSN_completion} for tensor completion problem.

\section{Conclusion}
In this paper, we establish several theoretical bounds for the problem of low-rank tensor recovery using random Gaussian measurements. For the nonconvex model \eqref{eqn:vector_non_convex}, we show that $\big{(}(2r)^K+2nrK+1 \big{)}$ measurements are sufficient to recover any $\mcb X_0 \in \mf T_r$ almost surely. We highlight that though the nonconvex recovery program is NP-hard in general, it does serve a baseline for evaluating tractable (convex) approaches. For the conventional convex surrogate sum-of-nuclear-norms (SNN) model \eqref{eqn:scalar_convex}, we prove a necessary condition that $\Omega(rn^{K-1})$ Gaussian measurements are required for reliable recovery. This lower bound is derived from our study on multi-structured object recovery under a very general setting, which can be applied to many scenarios. To narrow the apparent gap between the non-convex model and the SNN model, we unfold the tensor into a more balanced matrix while preserving its low-rank property, leading to our square-norm model \eqref{eqn:square_norm}. We prove that $O(r^{\lfloor \frac{K}{2} \rfloor}n^{\lceil \frac{K}{2} \rceil})$ measurements are sufficient to recover a tensor $\mcb X_0 \in \mf T_r$ with high probability. Though the theoretical results only pertain to Gaussian measurements, our simulation result for tensor completion also suggests that square-norm model outperforms the SNN model.

\begin{table}[h]
\centerline{
\begin{tabular}{|c|c|}
\hline
{ \bf Model } & {\bf sample complexity} \\
\hline \hline
non-convex  & $(2r)^K+2nrK+1$\\
\hline
SNN & $\Theta(rn^{K-1})$ \\
\hline
square-norm  &  $O(r^{\lfloor \frac{K}{2} \rfloor}n^{\lceil \frac{K}{2} \rceil})$\\
\hline
\end{tabular}}
\caption{\textbf{Summary of results derived in our paper.}} \label{tab:summary}
\end{table}

Compared with $\Omega(rn^{K-1})$ measurements required by the sum-of-nuclear-norms model, the sample complexity, $O(r^{\lfloor \frac{K}{2} \rfloor}n^{\lceil \frac{K}{2} \rceil})$, required by the square reshaping \eqref{eqn:square_norm}, is always within a constant of it, much better for small $r$ and $K\ge 4$. Although this is a significant improvement, compared with the nonconvex model \eqref{eqn:vector_non_convex}, the improved sample complexity achieved by square norm model is however still suboptimal. It remails an open problem to obtain near-optimal convex relaxations for all $K > 2$.

More broadly speaking, to recover objects with multiple structures, regularizing with a combination of individual structure-inducing norms is proven to be substantially suboptimal (Theorem \ref{thm:sim-struc} and also \cite{oymak2012simultaneously}). The resulting sample requirements tend to be much larger than the intrinsic degrees of freedom of the low-dimensional manifold that the structured signal lies in. Our square-norm model for the low-rank tensor recovery demonstrates the possibility that a better exploitation in those structures can significantly reduce this sample complexity. However, there are still no clear clues on how to intelligently utilize several simultaneous structures generally, and moreover how to design tractable method to recover multi-structured objects with near minimal number of measurements. These problems are definitely worth pursuing in future study and we hope that our work may also inspire researchers working in many other multi-structured recovery problems.

\section*{Acknowledgment}
It is a great pleasure to acknowledge conversations with Michael McCoy (Caltech), Ju Sun (Columbia), Han-wen Kuo (Columbia), Martin Lotz (Manchester), Zhiwei Qin (WalmartLabs). JW was supported by Columbia University startup funding and Office of Naval Research award N00014-13-1-0492.

\bibliographystyle{alpha}
\bibliography{Refs}

\newpage
\appendix

\section{Proofs for Section \ref{sec:nonconvex}}

\paragraph{Proof of Lemma \ref{lem:prob}.}

\begin{proof}
The arguments we used below are primarily adapted from \cite{eldar2011unicity}, where their interest is to establish the number of Gaussian measurements required to recover a low rank matrix by rank minimization.

Notice that every $\mcb D \in \mf S_{2r}$, and every $i$, $\innerprod{ \mcb{G}_i }{ \mcb D}$ is a standard Gaussian random variable, and so
\begin{equation}
\forall \,t > 0, \quad \prob{ \magnitude{ \innerprod{ \mcb{G}_i }{\mcb{D}} } < t }  < 2t \cdot \frac{1}{\sqrt{2\pi}}= t \sqrt{\frac{2}{\pi}}.
\end{equation}
Let $\mf N$ be an $\eps$-net for $\mf S_{2r}$ in terms of $\norm{\cdot}{F}$. Because the measurements are independent, for any fixed $\bar{\mcb{D}} \in \mf S_{2r}$,
\begin{equation}
\prob{ \norm{ \mc G[ \bar{\mcb{D}} ] }{\infty} < t } \;<\; \left(  t \sqrt{2/\pi} \right)^m.
\end{equation}
Moreover, for any $\mcb{D} \in \mf S_{2r}$, we have
\begin{eqnarray}
\norm{ \mc G[ \mcb{D} ] }{\infty} &\ge& \max_{\bar{\mcb{D}} \in \mf N} \set{ \; \norm{ \mc G[ \bar{\mcb{D}} ] }{\infty} - \norm{\mc G}{F\to \infty} \norm{\bar{\mcb{D}}- \mcb{D} }{F} \;} \\
&\ge& \min_{\bar{\mcb{D}} \in \mf N} \set{ \norm{\mc G[\bar{\mcb{D}}]}{\infty} } - \eps \norm{\mc G}{F \to \infty}.
\end{eqnarray}
Hence,
\begin{eqnarray}
\lefteqn{\prob{ \inf_{\mcb{D} \in \mf S_{2r}} \norm{\mc G[\mcb{D}]}{\infty} < \eps \log( 1/\eps ) } } \nonumber \\
&\le& \prob{ \min_{\mcb{D} \in \mf N} \norm{ \mc G[ \mcb{D} ] }{\infty}  < 2 \eps \log( 1/\eps ) } + \prob{ \norm{\mc G}{F \to \infty} > \log( 1/ \eps ) } \nonumber \\
&\le& \# \mf N \times \left( 2 \sqrt{2/\pi} \times \eps \log( 1/ \eps ) \right)^m + \prob{ \norm{\mc G}{F \to \infty} > \log( 1/ \eps ) } \nonumber \\
&\le& \beta^d (2\sqrt{2/\pi} )^m \eps^{{m-d}} \log( 1/\eps )^m +  \prob{ \norm{\mc G}{F \to \infty} > \log( 1/ \eps ) } \label{eqn: last_eqn} .
\end{eqnarray}
Since $m\ge d+1$, \eqref{eqn: last_eqn} goes to zero as $\eps \downto 0$. Hence, taking a sequence of decreasing $\eps$, we can show that $\prob{ \inf_{\mcb{D} \in \mf S_{2r}} \norm{ \mc G[\mcb{D}] }{\infty} = 0 } \le t$ for every positive $t$, establishing the result.
\end{proof}

\paragraph{Proof of Lemma \ref{lem:lipschitz}.}
\begin{proof}
This follows from the basic fact that for any tensor $\mcb{ X }$ and matrix $\mb U$ of compatible size,
\begin{equation}
\norm{ \mcb{X} \times_k \mb U }{F} \;\le\; \norm{\mb U}{} \norm{ \mcb{X} }{F},
\end{equation}
which can be established by direct calculation. Write
\begin{eqnarray*}
\lefteqn{\norm{ [[ \mcb{ C }; \mb U_1, \dots, \mb U_K ]] - [[ \mcb{ C }'; \mb U'_1, \dots, \mb U'_K ]] }{F} } \\
&\le& \norm{ [[ \mcb{ C }; \mb U_1, \dots, \mb U_K ]] - [[ \mcb{ C }'; \mb U_1, \dots, \mb U_K ]] }{F} \\ && \quad+ \quad \norm{ \sum_{i=1}^K [[ \mcb{ C }'; \mb U'_1, \dots, \mb U'_i, \mb U_{i+1} , \dots \mb U_k ]] -  [[ \mcb{ C }'; \mb U'_1, \dots, \mb U'_{i-1}, \mb U_{i} , \dots \mb U_K ]]  }{F} \\
&\le& \norm{ \mcb{ C} - \mcb{C}'}{F} + \sum_{i=1}^K \norm{\mb U_i - \mb U_i' },
\end{eqnarray*}
where the first inequality follows from triangle inequality and the second inequality follows from the fact that $\norm{\mcb{C}}{F} = 1$, $\norm{\mb U_j}{} = 1$, $\mb U_i^* \mb U_i = \mb I$ and ${\mb U_i'}^* \mb U_i' =  \mb I$.
\end{proof}

\paragraph{Proof of Lemma \ref{lem:covering-new}.}

\begin{proof}
The idea of this proof is to construct a net for each component of the Tucker decomposition and then combine those nets to form a {\em compound} net with the desired cardinality.

Denote $\mf C = \{\mcb C \in \reals^{2r\times 2r\times \cdots \times 2r} \mid \norm{\mcb C}{F}=1 \}$ and $\mc O =\{\mb U\in \reals^{n\times r} \mid \mb U^* \mb U =  \mb I\}$. Clearly, for any $\mcb{C} \in \mf C$, $\norm{\mcb C}{F}=1$, and for any $\mb U \in \mc O$, $\norm{\mb U}{}=1$. Thus by \cite[Prop. 4]{Ver2007lec} and \cite[Lemma 5.2]{vershynin2010introduction}, there exists an $\frac{\eps}{K+1}$-net $\mf C'$ covering $\mf C$ with respect to the Frobenius norm such that $\#\mf C'\le (\frac{3(K+1)}{\eps})^{(2r)^K}$, and there exists an $\frac{\eps}{K+1}$-net $\mc O'$ covering $\mc O$ with respect to the operator norm such that $\#\mc O'\le (\frac{3(K+1)}{\eps})^{2nr}$. Construct
$$\mf S_{2r}'=\{[[ \mcb{ C }'; \mb U'_1, \dots, \mb U'_K ]] \mid   \mcb{ C }' \in \mf C',\; \mb U'_i\in \mc O' \}.$$
Clearly $\#\mf S_{2r}'\le \left( \frac{ 3 (K+1) }{\eps} \right)^{ (2r)^K + 2 n r K} $. The rest is to show that $\mf S_{2r}'$ is indeed an $\eps$-net covering $\mf S_{2r}$ with respect to the Frobenius norm.

For any fixed $\mcb{D}=[[\mcb{C}; \mb U_1, \cdots, \mb U_K]]\in \mf S_{2r}$ where $\mcb{C} \in \mf C$ and $\mb U_i \in \mc O$, by our constructions above, there exist $\mcb C' \in \mf C'$ and $\mb U_i' \in \mc O'$ such that $\norm{\mcb{C}-\mcb{C}'}{F}\le \frac{ 3 (K+1) }{\eps}$ and $\norm{\mb U_i-\mb U'_i}{}\le \frac{ 3 (K+1) }{\eps}$. Then $\mcb{D}'=[[\mcb{C}'; \mb U'_1, \cdots, \mb U'_K]]\in \mf S'_{2r}$ is within $\eps$-distance from $\mcb{D}$, since by the triangle inequality derived in Lemma 2, we have
$$\norm{\mcb{D}-\mcb{D}'}{F}=\norm{ [[ \mcb{ C }; \mb U_1, \dots, \mb U_K ]] - [[ \mcb{ C }'; \mb U'_1, \dots, \mb U'_K ]] }{F}\le \norm{ \mcb{ C} - \mcb{C}'}{F} + \sum_{i=1}^K \norm{\mb U_i - \mb U_i' }{}\le \eps.$$
This completes the proof.
\end{proof}

\section{Proofs for Section \ref{sec:SNN}}

\paragraph{Proof of Corollary \ref{cor:kinematic}.}
\begin{proof}
Denote $\lambda = \delta(\mc C) - m$. Then following \cite[Thm. 7.1]{amelunxen2013living}, we have
\begin{eqnarray*}
\prob{ C \cap \mathrm{null}(\mc G)  = \set{ \mb 0 } } &\le & 4\exp \left( - \frac{ \lambda^2/8}{\min\{\delta(\mc C), \delta(\mc C^\circ)\}+\lambda} \right)\\
                         &\le& 4\exp \left( - \frac{ \lambda^2/8}{\delta(\mc C)+\lambda} \right)\\
                         &\le& 4 \exp \left( - \frac{ (\delta(C) - m)^2}{16 \delta(C) } \right).
\end{eqnarray*}
\end{proof}

\paragraph{Proof of Lemma \ref{lem:bd_cc}.}
\begin{proof}
Denote $\mbox{circ} (\mb e_n, \theta)$ as $\mbox{circ}_n(\theta)$, where $\mb e_n$ is the $n$th standard basis for $\reals^n$. Since $\delta \big{(} \mbox{circ} (\mb x_0, \theta) \big{)} = \delta \big{(} \mbox{circ} (\mb e_n, \theta) \big{)}$, it is sufficient to prove $\delta \big{(} \mbox{circ}_n(\theta) \big{)} \le n \sin^2 \theta +2$.

Let us first consider the case where $n$ is {\em even}. Define a discrete random variable $V$ supporting on $\{0,1,2,\cdots,n\}$ with probability mass function $\prob{V=k} = v_k$. Here $v_k$ denotes the $k$-th intrinsic volumes of $\mbox{circ}_n(\theta)$. As specified in \cite[Ex. 4.4.8]{amelunxen2011geometric}, we have
\begin{equation*}
v_k = \frac{1}{2} \binom{\frac{1}{2}(n-2)}{\frac{1}{2}(k-1)}\sin^{k-1}(\theta) \cos^{n-k-1}(\theta) \hspace{7mm}  \mbox{for} \; k = 1,2,\cdots,n-1.
\end{equation*}
From \cite[Prop. 5.11]{amelunxen2013living}, we know that
$$ \delta \big{(} \mbox{circ}_n(\theta) \big{)} = \expect{V} = \sum_{k=1}^n \prob {V \ge k}.$$
Moreover, by the interlacing result from \cite[Prop. 5.6]{amelunxen2013living} and the fact that $\prob{V \ge 2k} = \prob{V \ge 2k-1}-\prob{V=2k-1}$, we have
\begin{eqnarray*}
\begin{array}{lllll}
  \prob {V \ge 1} &\le & 2\prob {V=1} + 2\prob {V=3}+ \cdots + 2\prob {V=n-1}, \\
  \prob {V \ge 2} &\le &  \hspace{1.8mm} \prob {V=1} + 2\prob {V=3}+ \cdots + 2\prob {V=n-1}; \\
  \\
  \prob {V \ge 3} &\le & 2\prob {V=3} + 2\prob {V=5}+ \cdots + 2\prob {V=n-1},\\
  \prob {V \ge 4} &\le &  \hspace{1.8mm} \prob {V=3} + 2\prob {V=5}+ \cdots + 2\prob {V=n-1};\\
  \\
  \vdots &\vdots&\vdots\\
  \\
  \prob {V \ge n-1} &\le & 2\prob {V=n-1},\\
  \prob {V \ge n} &\le & \hspace{1.8mm} \prob {V=n-1}.\\
\end{array}
\end{eqnarray*}

Summing up the above inequalities, we have
\begin{eqnarray*}
 \expect{V} &=& \sum_{k=1}^n \prob {V \ge k} \\
            &\le&  \sum_{k=1,3,\cdots,n-1} 2(k-1)v_k+\sum_{k=1,3,\cdots,n-1}3v_k \\
            &\le& (n-2)\sin^2 \theta + \frac{3}{2} \sum_{k=0}^n  v_k\\
            &\le& (n-2)\sin^2 \theta + \frac{3}{2}\; = \; n\sin^2 \theta+2\cos ^2 \theta - \frac{1}{2},
\end{eqnarray*}
where the second last inequality follows the observations that $\sum_{k=1,3,\cdots,n-1} \frac{k-1}{2}\cdot (2v_k) = \expect{Bin(\frac{n-2}{2}, \sin^2 \theta)}$ and $\sum_{k=0}^n v_k \ge \sum_{k=1,3,\cdots,n-1}2v_k$ again by the interlacing result \cite[Prop. 5.6]{amelunxen2013living}.

Suppose $n$ is {\em odd}. Since the intersection of $\mbox{circ}_{n+1}(\theta)$ with any $n$-dimensional linear subspace containing $\mb e_{n+1}$ is an isometric image of $\mbox{circ}_n(\theta)$, by \cite[Prop. 4.1]{amelunxen2013living}, we have
$$
\delta(\mbox{circ}_n (\theta)) = \delta(\mbox{circ}_n (\theta)\times \{\mb 0\}) \le \delta(\mbox{circ}_{n+1} (\theta)) \le (n+1)\sin^2 \theta+2\cos ^2 \theta - \frac{1}{2} \le n\sin^2 \theta +\cos^2 \theta+\frac{1}{2}.
$$

Thus, taking both cases ($n$ is even and $n$ is odd) into consideration, we have $$\delta \big{(} \mbox{circ}_n(\theta) \big{)} \le n\sin^2 \theta  + \cos^2 \theta + \frac{1}{2} \le n\sin^2 \theta +2.$$

\end{proof}

\paragraph{Proof of Theorem \ref{thm:sim-struc}.}
\begin{proof}
Notice that for any fixed $m>0$, the function $f:\; t\to 4\exp\left(-\frac{(t-m)^2}{16t} \right)$ is decreasing for $t\ge m$. Then due to Corollary 4 and the fact that $\delta(\mc C)\ge \kappa-2 \ge m$, we have
\begin{eqnarray*}
\prob{ \mb x_0 \; \text{\rm is the unique optimal solution to \eqref{eqn:main_prob}} } &=& \prob{ C \cap \mathrm{null}(\mc G) = \set{ \mb 0 } } \\
   &\le&  4 \exp\left( - \frac{ (\delta(C) - m)^2}{16 \delta(C) } \right)\\
   &\le& 4 \exp\left( - \frac{( \kappa - m - 2)^2 }{16\, ( \kappa - 2 )} \right).
\end{eqnarray*}
\end{proof}

\section{Proofs for Section \ref{sec:square}}
\paragraph{Proof of Lemma \ref{lem: arrangement}.}

\begin{proof}
(1) By the definition of $\mcb{X}_{[j]}$, it is sufficient to prove that the vectorization of the right hand side of (4.2) equals $\mbox{vec}(\mcb{X}_{(1)})$.

Since $\mcb{X}=\sum_{i=1}^r \lambda_i \mb a_{i}^{(1)}\circ \mb a_{i}^{(2)}\circ \cdots \circ \mb a_{i}^{(K)}$, we have
\begin{eqnarray*}
\mbox{vec}(\mcb{X}_{(1)})&=& \mbox{vec}\big{(} \sum_{i=1}^r \lambda_i \mb a_i^{(1)}\circ (\mb a_i^{(K)}\otimes \mb a_i^{(K-1)}\otimes \cdots \otimes \mb a_i^{(2)})\big{)}\\
                       &=&  \sum_{i=1}^r \lambda_i \mbox{vec}\big{(} \mb a_i^{(1)} \circ (\mb a_i^{(K)}\otimes \mb a_i^{(K-1)}\otimes \cdots \otimes \mb a_i^{(2)}) \big{)}\\
                       &=& \sum_{i=1}^r \lambda_i (\mb a_i^{(K)}\otimes \mb a_i^{(K-1)}\otimes \cdots \otimes \mb a_i^{(2)}\otimes \mb a_i^{(1)}),
\end{eqnarray*}
where the last equality follows from the fact that $\mbox{vec}(\mb a \circ \mb b)= \mb b \otimes \mb a$.
Similarly, we can derive that the vectorization of the right hand side of (4.2),
\begin{eqnarray*}
  && \mbox{vec}(\sum_{i=1}^r \lambda_i(\mb a_i^{(j)}\otimes \mb a_i^{(j-1)}\otimes \cdots \otimes \mb a_i^{(1)})\circ (\mb a_{i}^{(K)} \otimes \mb a_{i}^{(K-1)} \cdots \otimes \mb a_{i}^{(j+1)}))  \\
  &=&   \sum_{i=1}^r \lambda_i \mbox{vec} \big{(}
  (\mb a_i^{(j)}\otimes \mb a_i^{(j-1)}\otimes \cdots \otimes \mb a_i^{(1)})\circ (\mb a_{i}^{(K)} \otimes \mb a_{i}^{(K-1)} \cdots \otimes \mb a_{i}^{(j+1)})
  \big{)} \\
  &=& \sum_{i=1}^r \lambda_i (\mb a_i^{(K)}\otimes \mb a_i^{(K-1)}\otimes \cdots \otimes \mb a_i^{(2)}\otimes \mb a_i^{(1)}) \\
  &=& \mbox{vec}(\mcb{X}_{(1)}).
\end{eqnarray*}
Thus, equation (4.2) is valid.
\\\\
(2) The above argument can be easily adapted to prove the second claim. Since $\mcb{X}=\mcb{C}\times_1 \mb U_1 \times_2 \mb U_2 \times_3 \cdots \times_K \mb U_K$, we have
\begin{eqnarray*}
  \mbox{vec}(\mcb{X}_{(1)}) &=& \mbox{vec} \bigg{(} \mb U_1 \; \mcb C_{(1)} \; (\mb U_K \otimes \mb U_{K-1} \otimes \cdots \otimes \mb U_2)^*  \bigg{)} \\
                            &=&  (\mb U_K \otimes \mb U_{K-1} \otimes \cdots \otimes \mb U_1) \; \mbox{vec} (\mcb C_{(1)}),
\end{eqnarray*}
where the last equality follows from the fact that $\mbox{vec}(\mb{ABC})= (\mb C^* \otimes \mb A) \mbox{vec}(\mb B)$. Similarly, we can derive that the vectorization of the right hand side of (4.3),
\begin{eqnarray*}
   && \mbox{vec} \bigg{(} (\mb U_j \otimes \mb U_{j-1} \otimes \cdots \otimes \mb U_1) \,\,\mcb{C}_{[j]}\,\, (\mb U_K \otimes \mb U_{K-1} \otimes \cdots \otimes \mb U_{j+1})^* \bigg{)}\\
   &=&  (\mb U_K \otimes \mb U_{K-1} \otimes \cdots \otimes \mb U_1)\; \mbox{vec} (\mcb C_{[j]})\\
   &=&  (\mb U_K \otimes \mb U_{K-1} \otimes \cdots \otimes \mb U_1) \; \mbox{vec} (\mcb C_{(1)})\\
   &=&   \mbox{vec}(\mcb{X}_{(1)}).
\end{eqnarray*}
Thus, equation (4.3) is valid.
\end{proof}

\section{Algorithms for Section \ref{sec:simulation}}
In this section, we will discuss in detail our implementation of accelerated linearized Bregman algorithm for the following problem:
\begin{eqnarray} \label{eqn:SSN}
\mbox{minimize}_{\mcb X} \quad \sum_{i=1}^K \|\tensorX_{(i)}\|_*  \quad \mbox{subject to} \quad \mc P_\Omega[\mcb X]=\mc P_\Omega[\mcb X_0].
\end{eqnarray}

By introducing auxiliary variable $\mcb W$ and splitting $\mcb X$ into $\mcb X_1,\; \mcb X_2,\; \cdots,\; \mcb X_K$, it can be easily verified that problem \eqref{eqn:SSN} is equivalent to
\begin{eqnarray}
\nonumber &\min_{(\{\mcb X_i\}, \mcb W)}& \sum_{i=1}^K \|(\mcb X_i)_{(i)}\|_* \\
          &\mbox{s.t. }& \mcb X_i = \mcb W, \quad  i=1,2,\cdots, K,   \label{eqn:SSN_trans}\\
\nonumber && \mc P_\Omega[\mcb W]=\mc P_\Omega[\mcb X_0],
\end{eqnarray}
whose objective function is now separable.

The accelerated linearized Bregman (ALB) algorithm, proposed in \cite{huang2013accelerated}, is an efficient first-order method designed for solving convex optimization problems with nonsmooth objective functions and linear constraints. It has been successfully applied to solve $\ell^1$ and nuclear norm minimization problems \cite{huang2013accelerated}. The ALB algorithm solves nonsmooth problem by firstly smoothing the objective function (e.g. adding a small $l_2$ perturbation), and then exploiting Nesterov's accelerated scheme \cite{nesterov1983method} to the dual problem, which can be verified to be unconstrained and Lipschitz differentiable. In Algorithm \ref{alg:ALB}, we describe our ALM algorithm adapted to problem \eqref{eqn:SSN_trans}. Algorithm \ref{alg:ALB} solves exactly the smoothed version of problem \eqref{eqn:SSN_trans}:
\begin{eqnarray}
\nonumber    &\min_{(\{\mcb X_i\}, \mcb W)}& \sum_{i=1}^K \left(\|(\mcb X_i)_{(i)}\|_*+ \frac{1}{2\mu}\|(\mcb X_i)_{(i)}\|_F^2\right)+ \frac{1}{2\mu}\|\mcb W\|_F^2\\
    &\mbox{s.t. }& \mcb X_i = \mcb W, \quad i=1,2,\cdots, K, \label{eqn:SSN_trans_smooth} \\
\nonumber    &            &  \mc P_\Omega[\mcb W]=\mc P_\Omega[\mcb X_0],
\end{eqnarray}
where we denote $\mcb Y_i$ as the dual variable for the constraint $\mcb X_i = \mcb W$ and denote $\mcb Z$ as the dual variable for the last constraint $\mc P_\Omega[\mcb W]=\mc P_\Omega[\mcb X_0]$. Since the objective function in \eqref{eqn:SSN_trans_smooth} is separable, each setup of the ALB algorithm is easy to solve as we can see from Algorithm \ref{alg:ALB} \footnote{ The Shrinkage operator in line 4 of Algorithm \ref{alg:ALB} performs the regular shrinkage on the singular values of the $i$th unfolding matrix of $\mcb Y_i^k$, i.e. $(\mcb Y_i^k)_{(i)}$, and then folds the resulting matrix back into tensor.}.

%

\begin{algorithm}[H]
\textbf{Initialization:} $\tensorY_i^0 = \tilde{\tensorY}_i^{0} = \mb 0$ for each $i\in [K]$, $\tensorZ^0=\tilde{\tensorZ}^{0}=\mb 0$, $\mu>0$, $\tau>0$, $t_0=1$\;

\For{$k=0,\; 1,\; 2, \; \cdots$}{
\For{$i=1,\; 2, \; \cdots,\; K$}{
$\tensorX_i^{k+1} = \mu \cdot \textrm{Shrinkage}(\tensorY_i^k, 1)$ \;
}
$\tensorW^{k+1} = \mu \cdot \left(\Proj_{\Omega}\left[\tensorZ^k\right] - \sum_i\tensorY_i^k\right)$\;
\For{$i=1,\; 2, \; \cdots,\; K$}{
$\tilde{\tensorY}_i^{k} = \tensorY_i^k - \tau \cdot \left(\tensorX_i^{k+1} - \tensorW^{k+1}\right)$ \;
}
$\tilde{\tensorZ}^{k} = \tensorZ^k - \tau \cdot \Proj_{\Omega}\left[\tensorW^{k+1} - \tensorX_0\right]$\;
$t_{k+1} = \frac{1+\sqrt{1+4t_k^2}}{2}$\;
\For{$i=1,\; 2, \; \cdots,\; K$}{
$\tensorY^{k+1}_i = \tilde{\tensorY}_i^k + \frac{t_k-1}{t_{k+1}}\left(\tilde{\tensorY}^k_i - \tilde{\tensorY}^{k-1}_i\right)$ \;
}
$\tensorZ^{k+1} = \tilde{\tensorZ}^k + \frac{t_k-1}{t_{k+1}}\left(\tilde{\tensorZ}^k - \tilde{\tensorZ}^{k-1}\right)$\;
}
\caption{accelerated linearized Bregman algorithm for SNN model \eqref{eqn:SSN}}
\label{alg:ALB}
\end{algorithm}

For our numerical experiment ($K=4$), we choose smoothing parameter $\mu = 50\|\tensorX_0\|_F$ and step size $\tau = \frac{1}{5 \mu}$. Empirically, we observe that larger values of $\mu$ do not result in a better recovery performance. This is consistent with the theoretical results established in \cite{lai2012augmented, zhang2011strongly}.

\end{document}